\documentclass[11pt]{article}

\usepackage{arxiv}
\usepackage[utf8]{inputenc}
\usepackage[T1]{fontenc}
\usepackage{lmodern}
\usepackage[colorlinks=true, linkcolor=snsblue, citecolor=snsgreen, urlcolor=snsred]{hyperref}
\usepackage{url}
\usepackage{booktabs}
\usepackage{amsmath}
\usepackage{amssymb}
\usepackage{amsfonts}
\usepackage[normalem]{ulem}
\usepackage{nicefrac}
\usepackage{microtype}
\usepackage{natbib}
\usepackage{siunitx}
\usepackage{mathtools}
\usepackage{algorithm}
\usepackage{algpseudocode}
\usepackage{csquotes}
\usepackage[nameinlink]{cleveref}

\usepackage{xcolor}
\definecolor{snsblue}{RGB}{76, 114, 176}
\definecolor{snsgreen}{RGB}{85, 168, 104}
\definecolor{snsred}{RGB}{196, 78, 82}
\definecolor{snspurple}{RGB}{129, 114, 178}
\definecolor{snsyellow}{RGB}{204, 185, 116}
\definecolor{snscyan}{RGB}{100, 181, 205}

\title{Batched NN 4D-Var: a performant hybrid model framework for the Integrated Forecasting System}

\author{
Alban Farchi\\
ECMWF\\
Bonn, Germany\\
\texttt{alban.farchi@ecmwf.int}\\
\And
Marcin Chrust\\
ECMWF\\
Reading, United Kingdom\\
\And
Massimo Bonavita\\
ECMWF\\
Reading, United Kingdom\\
\And
Marc Bocquet\\
CEREA, ENPC, EDF R\&D\\
Institut Polytechnique de Paris\\
\^Ile--de--France, France\\
\And
Wei Pan\\
ECMWF\\
Reading, United Kingdom\\
\And
Patrick Laloyaux\\
ECMWF\\
Reading, United Kingdom
}

\begin{document}

\maketitle

\begin{abstract}
Hybrid forecasting systems combining machine learning and numerical weather prediction offer a promising route toward improved forecast skill while preserving physical consistency. In this study, we investigate online learning of state-dependent model-error corrections within the Integrated Forecasting System using four-dimensional variational data assimilation (4D-Var). Building on previous work, we represent model error using a neural network embedded within the forecast model and trained from observations through the data assimilation cycle.

First, we introduce a targeted variant of neural-network 4D-Var in which only the neural-network parameters are estimated, while the atmospheric state is provided by a precomputed analysis. We then propose a batched extension that processes assimilation windows in random order and averages parameter increments over multiple windows, drawing inspiration from mini-batch stochastic optimization.

Experiments are conducted with the Integrated Forecasting System of the European Centre for Medium-Range Weather Forecasts at an intermediate resolution. Sequential online learning substantially improves forecasts within the training period, with error reductions exceeding \qty{10}{\percent} in some cases. However, these improvements fail to generalize to unseen data. In contrast, the proposed batched formulation produces hybrid models that consistently improve forecast skill in out-of-sample experiments. Larger batch sizes accelerate convergence and improve robustness by reducing the noise of parameter updates. Diagnostics further show that the online-trained corrections preserve realistic forecast variability and do not improve skill through artificial smoothing. These results demonstrate that the formulation of the online learning problem is critical for generalisation and establish batched online training as a promising framework for hybrid machine-learning/data-assimilation systems.
\end{abstract}


\section{Introduction}

Over the past few years, machine‑learning‑based weather prediction (MLWP) has transitioned from an emerging to an established and rapidly evolving area of research. Global MLWP systems trained on decades of global reanalysis data are now able to match and even surpass state‑of‑the‑art physics-based numerical weather prediction (NWP) models across a wide range of forecast lead times and large-scale verification metrics, while requiring only a fraction of the computational cost. This technological breakthrough shows that artificial intelligence in general, and machine learning (ML) in particular, introduced a structural shift in how weather prediction systems may be designed in the future. This transition has already moved beyond the research stage, with ECMWF operationalising its Artificial Intelligence Forecasting System (AIFS) in 2025 in both deterministic and ensemble configurations.

The first wave of MLWP models \citep{pathak-2022, bi-2023, lam-2023, lang-2024} were by construction deterministic: these models were trained using supervised learning with the ERA5 reanalysis \citep{hersbach-2020} produced by the European Centre for Medium-Range Weather Forecasts (ECMWF). These models have demonstrated remarkable forecast skill across a broad range of verification measures, including root-mean-squared error (RMSE), anomaly correlation, and mean bias, as well as for high-impact phenomena such as tropical cyclone tracks and weather extremes. Despite these successes, some limitations of this first generation of MLWP models have also been identified \citep{bocquet-2023, bonavita-2023, pasquini-2026}. First, they show a progressive smoothing of the predictions over the forecast lead time. This issue is now well understood as a consequence of using mean-squared errors (MSE) as loss function: the MLWP models tend to smooth their predictions in order to circumvent the so-called double penalty effect \citep{lang-2024b}. Second, because physical and dynamical constraints are not explicitly imposed during training, maintaining all relevant dynamical balances can remain challenging, despite the excellent forecast skill demonstrated by these models. This aspect becomes particularly important when MLWP models are used within cycled data assimilation systems, where the forecast model is repeatedly coupled to the assimilation process and small inconsistencies may accumulate over successive cycles \citep{slivinski-2025, tian-2026}.

Consequently, a second wave of MLWP models has emerged, relying on advances in generative ML \citep{price-2024, finn-2024, lang-2024b, finn-2025}. Diffusion models and related probabilistic generative techniques explicitly represent forecast uncertainty and are trained to sample from the full conditional distribution of future atmospheric states. By construction, these models alleviate the symptoms of the smoothing problem associated with deterministic MSE‑based training by injecting phase-blind noise at the missing small scales, resulting in more realistic-looking power spectra of the relevant dynamical variables. While generative MLWP models mark a significant conceptual advance, their integration into existing Data Assimilation (DA) infrastructures remains an open research question \citep{slivinski-2025, tian-2026}.

An alternative and complementary route toward effectively leveraging the capabilities of machine learning for weather prediction is hybrid modelling, in which machine learning components are embedded within, or coupled to, traditional physics‑based models \citep{levine-2022}. Rather than replacing NWP systems, hybrid approaches seek to augment them, preserving the capabilities of the physical model while benefiting from data-driven information. A wide range of hybrid strategies has been proposed \citep[see][and references therein]{farchi-2025}, but two major directions have recently drawn more attention: ML‑based model error correction \citep{farchi-2025} and spectral nudging \citep{husain-2025}.

In spectral nudging methods \citep{husain-2025, polichtchouk-2026}, the large‑scale dynamics of a forecasting system are nudged towards a pre-computed MLWP forecast trajectory, while the small scales are left untouched. These methods have only recently appeared in the literature and their appeal lies in their simplicity and ease of deployment: they can often be implemented with minimal changes to existing forecast systems.

The other main direction, ML‑based model error correction is challenging to implement in an operational system but offers the prospect of a fully consistent physical-statistical framework, in which a neural network is explicitly trained to represent the model error of the underlying physical model \citep{bonavita-2020, wikner-2020, farchi-2021, farchi-2021b, brajard-2020b, chen-2022}. The resulting model of model error can then be used to compensate for structural biases and deficiencies of the physical model during a forecast. The present study builds on the recent work of \citet{farchi-2023, farchi-2025}, hereafter referred to as F23 and F25, who introduced a new hybrid data assimilation framework for online model error correction. In their approach, a neural network is trained online using a variant of weak‑constraint four‑dimensional variational (4D-Var) assimilation system, which they called neural network (NN) 4D-Var. While NN 4D‑Var represents a significant conceptual advance, it also exhibits important limitations, most notably a poor generalisation to unseen data as will be illustrated here. 

In this article, we propose a new batched variant of the NN 4D‑Var framework designed to address these shortcomings. By reformulating the training strategy to exploit random batches of assimilation windows in parallel, we aim to improve the statistical robustness and generalisation capability of the neural network, thereby increasing the forecast accuracy of the hybrid model. The article is structured as follows: \cref{sec:methodology} presents the methodology, explaining in particular how and why we chose to move from offline- to batched training. In \cref{sec:application}, the new method is applied to the ECMWF Integrated Forecasting System \citep[IFS,][]{bonavita-2017}. The results are discussed in \cref{sec:discussion} and conclusions are given in \cref{sec:conclusions}.

\section{From offline to batch training}
\label{sec:methodology}

In this section, we present the methodological foundations of the present study. We first introduce the hybrid model formulation that underpins all subsequent experiments. We then describe the offline training strategy based on analysis increments, which serves as a baseline approach. Next, we recall the original NN 4D-Var framework proposed by \citet{farchi-2021b}, which provides a first formulation of online training for neural network-based model error correction. Finally, we introduce two modifications of the NN 4D-Var method aimed at (i) steering the assimilation process towards a more effective estimation of model parameters and (ii) improving the generalisation properties of the resulting hybrid model. These two modifications constitute the main methodological contributions of the present work.

\subsection{Hybrid model formulation}

Physics-based forecast models used in NWP rely on the numerical integration of laws governing atmospheric motion and physics. These laws are typically expressed as systems of ordinary or partial differential equations (ODEs or PDEs), which describe the temporal evolution of the atmospheric state. With $\mathbf{x} \in \mathbb{R}^{N_{\mathsf{x}}}$ denoting the discretised state of the system, representing the atmospheric variables on a numerical grid, the model dynamics can be written in the generic form
\begin{equation}
    \frac{\mathrm{d}\mathbf{x}}{\mathrm{d}t} = \boldsymbol{\phi}(\mathbf{x}),
\end{equation}
where $\boldsymbol{\phi}$ are the model tendencies.

To obtain a practical forecast model, these tendencies are integrated forward in time using a numerical scheme:
\begin{equation}
    \boldsymbol{\mathcal{I}}_k : \mathbf{x}_k \mapsto \mathbf{x}_{k+\delta t},
\end{equation}
which advances the state of the system from $t_{k}$ to $t_{k+\delta t}$ by one time step $\delta t$. A full forecast is obtained by composing several integration steps:
\begin{equation}
    \boldsymbol{\mathcal{M}}_{l:k} : \mathbf{x}_k \mapsto \mathbf{x}_l = \boldsymbol{\mathcal{I}}_{l-\delta t} \circ \cdots \circ \boldsymbol{\mathcal{I}}_k(\mathbf{x}_k),
\end{equation}
where $\boldsymbol{\mathcal{M}}_{l:k}$ is the model resolvent from $t_{k}$ to $t_{l}$, usually simply called the forecast model.

As explained in \citet{farchi-2021b}, there exist multiple strategies to correct or augment a physical forecast model using ML. These strategies differ in where and how the data-driven component is introduced, ranging from direct modifications of the model tendencies to corrections of the model resolvent. In this work, we consider an intermediate approach in which an additive correction is applied after each integration step. Specifically, we define a hybrid integration operator $\boldsymbol{\mathcal{I}}^{\mathsf{h}}_{k}(\mathbf{p})$ as
\begin{equation}
    \boldsymbol{\mathcal{I}}^{\mathsf{h}}_{k}(\mathbf{p}) : \mathbf{x}_k \mapsto \boldsymbol{\mathcal{I}}_k(\mathbf{x}_k) + \boldsymbol{\mathcal{F}}(\mathbf{p}, \mathbf{x}_k),    
\end{equation}
where $\boldsymbol{\mathcal{F}}(\mathbf{p}, \cdot)$ is a correction term, parametrised by a set or parameters $\mathbf{p}$. The corresponding hybrid forecast model is then obtained by composing the hybrid integration operator over successive time steps:
\begin{equation}
\boldsymbol{\mathcal{M}}^{\mathsf{h}}_{l:k} : (\mathbf{p}, \mathbf{x}_k) \mapsto \mathbf{x}_l = \boldsymbol{\mathcal{I}}^{\mathsf{h}}_{l-\delta t}(\mathbf{p}) \circ \cdots \circ \boldsymbol{\mathcal{I}}^{\mathsf{h}}_{k}(\mathbf{p})(\mathbf{x}_k).
\end{equation}
Because the correction term is applied at each time step and directly modifies the discrete-time tendencies, it is conceptually closer to tendency correction, while still operating at the level of the numerical integration scheme rather than the continuous-time equations. This formulation provides a natural framework for coupling machine learning-based model error corrections with existing NWP models in a dynamically consistent manner.

It is important to note that all online\footnote{Note that the difference between offline and online training is explained in \cref{ssec:nomenclature-offline-online}.} training methods presented in this work do not depend on the specific form of the model error correction, but only on the fact that the forecast model is parametrised. In particular, they could even be used to train fully data-driven models. By contrast, the offline training strategy introduced in the following section does rely on the correction being additive. Following F23 and F25, in our experimental setup, offline training is used exclusively as a pre-training step, providing initial parameter values before the online training experiments are started. Furthermore, $\boldsymbol{\mathcal{F}}$ is implemented via a neural network, with $\mathbf{p}$ being its weights and biases.

\subsection{Offline training from analysis increments}
\label{ssec:methodology-offline}

In a well-calibrated data assimilation (DA) system, analysis increments can be interpreted as a proxy for forecast errors that develop over one DA window \citep{farchi-2021}. Offline training strategies exploit this property by training neural networks to predict model error using increments extracted from a large set of analyses or from a reanalysis dataset. Typically, a neural network is trained in a prediction mode, where
\begin{itemize}
    \item the input is the analysis at the beginning of a window, $\mathbf{x}^{\mathsf{a}}(t)$;
    \item the output is the analysis increment at the beginning of the next window, $\delta \mathbf{x}^{\mathsf{a}}(t+L\times\Delta T)$;
\end{itemize}
where $L\times\Delta T$ is the length of the assimilation window. In our experiments, the neural network output is then multiplied by $\delta t / (L\times\Delta T)$, where $\delta t$ is the size of the time step, to rescale the increment from an error accumulated over one assimilation window to an effective per–time-step correction, under the assumption of linear error growth in time. This rescaling allows the trained neural network to be used as an additive correction within the hybrid model framework introduced in the previous section and to perform forecast experiments.

This approach is referred to as \emph{offline training} because it does not explicitly require the physical forecast model: it relies solely on precomputed analyses and forecasts stored in a dataset. However, it has been shown that the rescaling assumption constitutes the main limitation of this strategy. Due to nonlinear error growth, forecast errors accumulated over an assimilation window can significantly differ from the instantaneous model error acting at each time step, as required in the hybrid model. This limitation motivates the development of \emph{online} training approaches, in which the interaction between the physical model and the data-driven correction is explicitly taken into account during the training process \citep{farchi-2021b}.

\subsection{Online training with NN 4D-Var}
\label{ssec:methodology-nn-4d-var}

While offline training strategies are relatively straightforward to implement, online training is notoriously more challenging. In order to account for interactions between the physical forecast model and the data-driven correction during the training process, one must in principle differentiate through the forecast model itself. This requires access to tangent-linear and adjoint operators of the physical model, which are rarely available for modern global NWP systems and are costly to develop and maintain.

In practice, most online training approaches for model error correction are formulated within existing DA frameworks. A common strategy is to augment the control vector to include both the model state and a set of model parameters, which are then estimated simultaneously during the assimilation process. In the present study, we build upon the NN 4D-Var method introduced by \citet{farchi-2021b}. This method can be seen as a parametrised variant of weak-constraint 4D-Var, in which model error is implicitly represented through a neural network whose parameters are estimated alongside the initial state. In the original formulation, the observations $\mathbf{y}_{k}, k\in\mathbb{Z}$ are assimilated by windows of length $L\times\Delta T$. For the window starting at $t_0$, the analysis is obtained by minimising the cost function
\begin{equation}
    \label{eq:nn-4d-var-cost}
    \mathcal{J}(\mathbf{p}, \mathbf{x}_{0}) = \frac{1}{2}\left\|\mathbf{x}_{0}-\mathbf{x}^{\mathsf{b}}_{0}\right\|^{2}_{\mathbf{B}^{-1}} + \frac{1}{2}\left\|\mathbf{p}-\mathbf{p}^{\mathsf{b}}\right\|^{2}_{\mathbf{P}^{-1}} + \frac{1}{2}\sum_{k=0}^{L-1}\left\|\mathbf{y}_{k} - \boldsymbol{\mathcal{H}}_{k}\circ \boldsymbol{\mathcal{M}}^{\mathsf{h}}_{k:0}\left(\mathbf{p}, \mathbf{x}_{0}\right)\right\|^{2}_{\mathbf{R}^{-1}_{k}},
\end{equation}
where $\left\|\mathbf{v}\right\|_{\mathbf{M}}=\sqrt{\mathbf{v}^{\top}\mathbf{Mv}}$ is the Mahalanobis, or weighted $L^{2}$, norm. Under standard Gaussian error assumptions, this cost function corresponds to the negative log-likelihood of the joint posterior distribution of the initial state $\mathbf{x}_0$ and the model parameters $\mathbf{p}$ given the observations $\mathbf{y}_{k}, k=0, \ldots, L-1$.

The first term represents the background error, where $\mathbf{x}^{\mathsf{b}}_{0}$ is the background initial state and $\mathbf{B}$ the associated background error covariance matrix. The second term plays an analogous role for the model parameters, with $\mathbf{p}^{\mathsf{b}}$ denoting the background parameter and $\mathbf{P}$ the parameter error covariance matrix. The third term represents the observation errors throughout the assimilation window, where $\boldsymbol{\mathcal{H}}_k$ is the observation operator and $\mathbf{R}_k$ the observation error covariance matrix at time $t_k$.

Minimising $\mathcal{J}$ yields an analysis for both the initial state $\mathbf{x}^{\mathsf{a}}_0$ and the model parameters $\mathbf{p}^{\mathsf{a}}$. After each assimilation cycle, the state analysis $\mathbf{x}^{\mathsf{a}}_0$ is propagated to the next window using the hybrid model $\boldsymbol{\mathcal{M}}^{\mathsf{h}}_{L:0}$. The parameter analysis is carried forward in time unchanged, corresponding to the persistence model for $\mathbf{p}^{\mathsf{a}}$.

\subsection{Tailoring NN 4D-Var to focus on model parameters}
\label{ssec:methodology-targ-nn-4d-var}

Joint estimation of the initial state and model parameters within a DA framework possesses several appealing theoretical properties. In particular, when correctly specified, it can in principle yield a more accurate and dynamically consistent estimate by exploiting the mutual dependencies between initial conditions and model error. However, joint assimilation is also significantly more difficult to implement in practice, and its success critically depends on the ability of the assimilation system to distinguish errors in the initial state from errors in the model parameters.

This separation is largely controlled by the relative weighting of the background error covariance matrices associated with the initial state ($\mathbf{B}$) and with the model parameters ($\mathbf{P}$). While background error covariances for the state variables can be estimated and tuned using well-established techniques, the calibration of prior error statistics for model parameters remains largely empirical. Poorly calibrated covariances may lead to an undesirable aliasing of initial state errors into parameter increments \citep{laloyaux-2020}, thereby degrading the quality of both the parameter estimation and the resulting hybrid model.

For these reasons, we propose a simplified variant of NN 4D-Var without joint estimation. Instead, we assume that the observations have already been assimilated using a standard DA method to estimate the initial state only, yielding a precomputed analysis $\mathbf{x}^{\mathsf{pa}}_{0}$, which is frozen in the subsequent parameter estimation step. Starting from this precomputed analysis, we re-assimilate the observations in order to estimate the model parameters $\mathbf{p}$ only, by minimising the following cost function:
\begin{equation}
\label{eq:targ-nn-4d-var-cost}
    \mathcal{J}(\mathbf{p}) = \frac{1}{2}\left\|\mathbf{p}-\mathbf{p}^{\mathsf{b}}\right\|^{2}_{\mathbf{P}^{-1}} + \frac{1}{2}\sum_{k=0}^{L-1}\left\|\mathbf{y}_{k} - \boldsymbol{\mathcal{H}}_{k}\circ \boldsymbol{\mathcal{M}}^{\mathsf{h}}_{k:0}\left(\mathbf{p}, \mathbf{x}^{\mathsf{pa}}_{0}\right)\right\|^{2}_{\mathbf{R}^{-1}_{k}}.
\end{equation}
As in the original NN 4D-Var formulation, the parameter analysis is carried forward in time unchanged from one assimilation window to the next. This method is highly general and does not depend on how the precomputed analysis has been obtained. However, for the sake of clarity, we can assume for now that $\mathbf{x}^{\mathsf{pa}}_{0}$ is coming from a reanalysis dataset or directly from the operational archive, as will be the case in the numerical experiments shown in \cref{sec:application}.

This cost function is equivalent to that of NN 4D-Var, \cref{eq:nn-4d-var-cost}, in the limiting case where the background error covariance for the initial state tends to zero, $\mathbf{B}\to\mathbf{0}$, and the background state $\mathbf{x}^{\mathsf{b}}_{0}$ coincides with the precomputed analysis $\mathbf{x}^{\mathsf{pa}}_{0}$. In other words, this formulation corresponds to an NN 4D-Var system in which all degrees of freedom associated with the initial state have been frozen in the control vector.

\begin{figure}
    \begin{center}
        \includegraphics[width=0.99\linewidth]{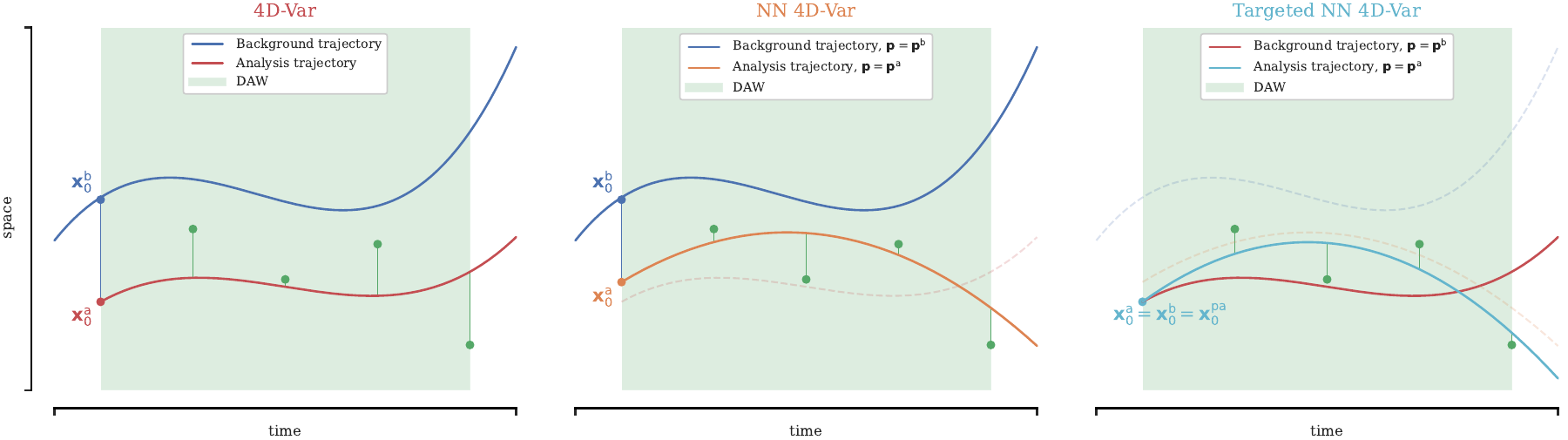} 
    \end{center}
    \caption{Illustration of the differences between 4D-Var (left), NN 4D-Var (middle), and targeted NN 4D-Var (right). In 4D-Var, the goal is to estimate the initial state $\mathbf{x}_{0}$ only. NN 4D-Var provides a joint estimation of both initial state $\mathbf{x}_{0}$ and model parameters $\mathbf{p}$, while targeted NN 4D-Var focus on model parameters $\mathbf{p}$ only and relies on a pre-computed analysis $\mathbf{x}^{\mathsf{pa}}_{0}$ for the initial state.}
    \label{fig:4d-var-variants}
\end{figure}

We refer to this approach as \emph{targeted NN 4D-Var}, as it essentially corresponds to the NN 4D-Var framework targeted exclusively at the estimation of model parameters. By construction, this design prevents the assimilation from compensating for model error through spurious state adjustments, and instead forces the learning process to focus directly on the neural network-based model error correction. On the other hand, one should keep in mind that, with this method, observations are assimilated twice: once to estimate the state and once to estimate the model parameters. This raises questions about whether the new method is truly Bayesian. While we cannot fully answer this question, we propose a justification of this new method in the context of the iterative framework of \citet{bocquet-2020, brajard-2020} in \cref{ssec:discussion-coordinate-descent}. Finally, a schematic illustration of the progression from classical 4D-Var to NN 4D-Var, and finally to the targeted NN 4D-Var framework introduced here, is provided in \cref{fig:4d-var-variants}.

\subsection{Batching NN 4D-Var to enhance generalisation}
\label{ssec:methodology-batched-nn-4d-var}

In the original NN 4D-Var formulation, assimilation windows must be processed in strict chronological order. This constraint stems from the cycled nature of the method, in which the background state for each window is provided by the analysis of the previous one, as in any standard sequential DA system. While this scheduling is well suited to state estimation, it imposes strong limitations on the training of model parameters.

In practice, NWP models are not autonomous, and model error often exhibits pronounced temporal variability, for instance seasonal or regime-dependent components. In such a context, a strictly sequential learning process is not ideal for identifying general model error patterns. Indeed, learning in a sequential framework requires the system to retain information from a large number of previous cycles, which in practice is achieved by specifying a small background error covariance matrix for the model parameters, $\mathbf{P}$. While this enhances long-term memory, it comes at the cost of reduced adaptability and slow learning, particularly when the model error statistics vary in time.

These considerations motivate the exploration of non-sequential training strategies, such as assimilating windows in a random order. In the targeted NN 4D-Var framework introduced above, this is feasible because the analysis of the initial state is assumed to be precomputed and fixed. As a result, assimilation windows are statistically independent from each other, and can therefore be processed in any order without violating the consistency of the method.

\begin{figure}
    \centering
    \includegraphics[width=0.99\linewidth]{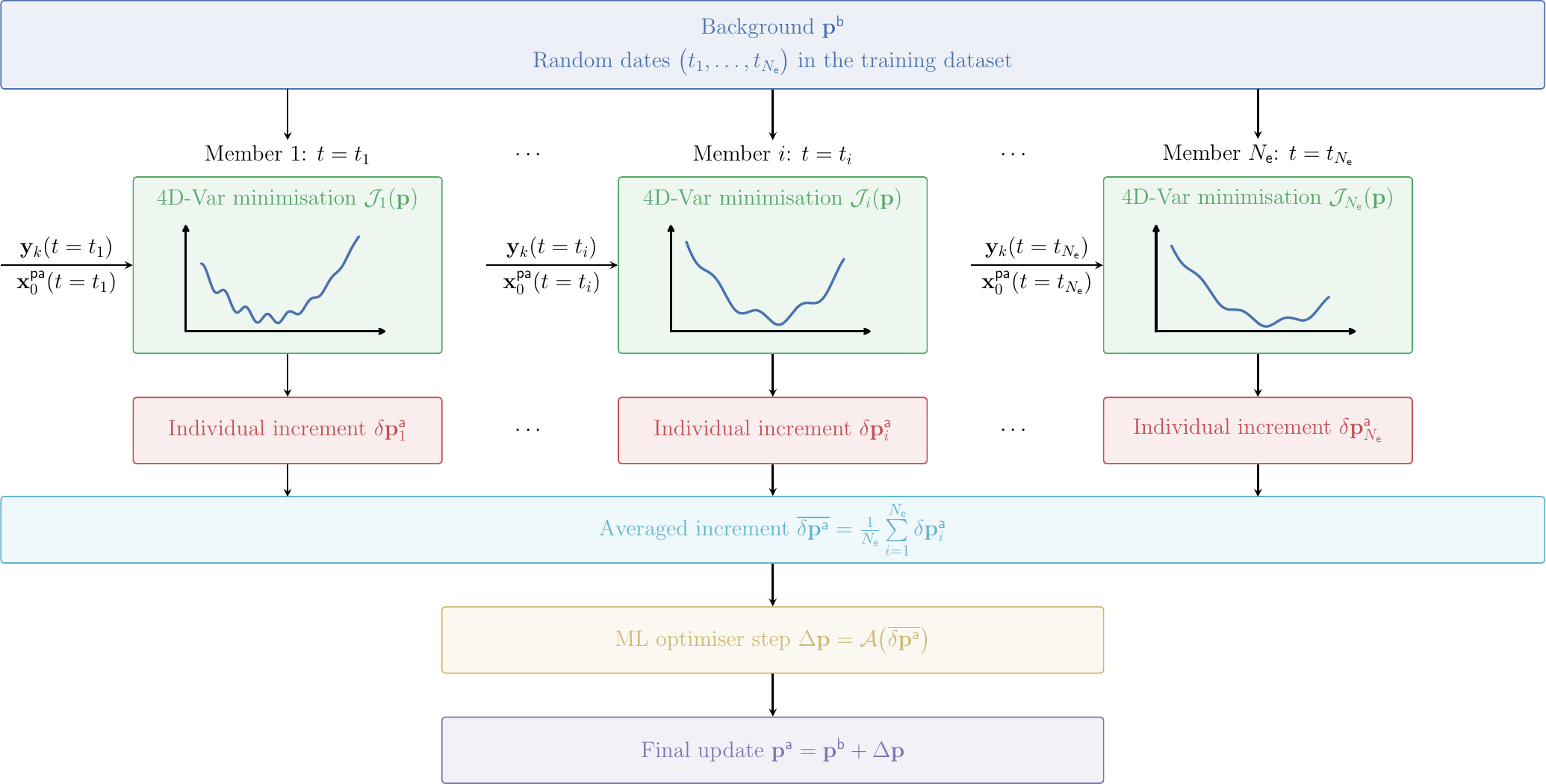}
    \caption{Illustration of an assimilation cycle in the batched targeted NN 4D-Var method. The cycle starts from a background for model parameters $\mathbf{p}^{\mathsf{b}}$. Random dates are sampled from the training dataset and assigned to a member in the batch. Each member then solves its own targeted NN 4D-Var problem, yielding individual parameter increments $\delta\mathbf{p}^{\mathsf{a}}_{i}$, which are then averaged in a synchronisation step. An optional ML optimiser step follows (see \cref{ssec:discussion-similarities-sgd}) before the parameter analysis $\mathbf{p}^{\mathsf{a}}$ gets computed.}
    \label{fig:batch-variant}
\end{figure}

Furthermore, taking inspiration from machine learning techniques, in which mini-batch strategies are standard, we propose a batched extension of targeted NN 4D-Var. In this \emph{batched NN 4D-Var} approach, each assimilation cycle processes a batch of $N_{\mathsf{e}}$ assimilation windows, randomly selected from a training dataset. Each window is assimilated independently using the targeted NN 4D-Var formulation, producing an individual analysis increment for the model parameters, denoted $\delta\mathbf{p}^{\mathsf{a}}_{i}$ for the $i$-th window. In a final synchronisation step, these increments are averaged and applied to the background parameters:
\begin{equation}
    \mathbf{p}^{\mathsf{a}} = \mathbf{p}^{\mathsf{b}} + \frac{1}{N_{\mathsf{e}}}\sum_{i=1}^{N_{\mathsf{e}}}\delta\mathbf{p}^{\mathsf{a}}_{i}.
\end{equation}
A full assimilation cycle of the batched NN 4D-Var method is illustrated schematically in \cref{fig:batch-variant}. By averaging the parameter increments over a mini-batch of assimilation windows, the intrinsic noise associated with individual increments is reduced, leading to a more robust update of the model parameters. In addition, processing multiple windows in parallel substantially increases the rate at which observations are assimilated, making it possible to expose the model to $N_{\mathsf{e}}$ times more observational information for a fixed number of optimisation steps. Together, these effects are expected to enhance the ability of the system to learn generalisable model error representations.

\subsection{Nomenclature: offline versus online learning}
\label{ssec:nomenclature-offline-online}

To conclude this section, let us clarify the terminology used to distinguish between online and offline learning, as different definitions coexist in the literature.

In the geoscientific data assimilation community, offline learning typically refers to methods in which the training is performed once a complete dataset is available. By contrast, online learning denotes approaches in which the model is updated progressively as new batches of observations become available, typically within a cycled data assimilation framework. In the machine learning community, the distinction is of a different nature. Online learning refers to training procedures that explicitly involve the dynamical system (in the present context, the physics-based forecast model) within the training loop. Conversely, offline learning corresponds to situations in which the data-driven model is trained independently of the dynamical system, using a precomputed dataset of inputs and targets.

Until now, this distinction was largely inconsequential, as previously proposed training strategies were either offline according to both definitions (e.g., training on precomputed analysis increments) or online according to both definitions (e.g., joint state-parameter estimation within NN 4D-Var). However, the targeted NN 4D-Var and batched NN 4D-Var formulations introduced in this work do not fit neatly into this dichotomy.

On the one hand, these methods rely on a precomputed analysis and allow assimilation windows to be processed in a random order, which would qualify them as offline methods under the geoscientific definition. On the other hand, the parameter updates are obtained through a data assimilation procedure that explicitly involves the hybrid forecast model, meaning that the physics-based model is part of the training process. According to the machine learning definition, these methods should therefore be considered online. In the remainder of this paper, we adopt the machine learning convention and refer to these approaches as online training methods. Nevertheless, it should be kept in mind that, under the geoscientific convention, they exhibit characteristics that are typically associated with offline learning.

\section{Application to the IFS}
\label{sec:application}

\subsection{Technical experimental details}
\label{ssec:application-technical-detauls}

All experiments presented in this study are conducted with a DA configuration similar to that of F25. The experiments are carried out using the Object-Oriented Prediction System (OOPS) developed at ECMWF, but rely on a more recent version of the forecast model, namely the IFS cycle 49r1 (instead of 48r1 in F25). We employ a standard research configuration of incremental 4D-Var with three outer loops. The nonlinear forecast model is integrated at horizontal resolution TCo399, while the inner-loop resolutions are set to TL95, TL159, and TL255.

The neural-network-based model error correction used throughout this work is identical to the one introduced in F25 and inspired from \citet{bonavita-2020}. It follows a column-based architecture and provides corrections for four atmospheric variables: temperature (t), logarithm of surface pressure (lnsp), vorticity (vo), and divergence (d). In addition to the model state, date, time, latitude, and longitude are used as extra predictors. The network consists of four hidden layers with hyperbolic tangent activation functions, and employs variable- and level-dependent normalisation factors. The total number of trainable parameters is \num{1214876}. In the online experiments, the neural network is implemented using the Fortran neural network library \citep[FNN,][]{fnn-2022}, as in the original study. Finally note that, throughout the entire study, in the online experiments the correction is masked in the tropics (between latitudes \qty{-30}{\degree} and \qty{+30}{\degree}), as it was found to degrade the forecast skill in this region. Re-introducing the correction in the tropics is left for future work, as explained in the conclusions, in \cref{sec:conclusions}.

The datasets used for training and evaluation are separated in time to assess the generalisation capabilities of the hybrid model. Offline training experiments are performed using DA increments and analyses over 2017--2022. Online training experiments use observations from the year 2024 only. Finally, all forecast experiments used to evaluate the performance of the trained hybrid models are conducted over the year 2025, ensuring a clear temporal separation between training and validation data. In the latter case, forecasts are initialised from the operational analyses at 00:00 UTC, starting on 2025-01-01 and ending on 2025-12-31. Forecasts are launched every \qty{48}{\hour}, resulting in a total of \num{183} forecasts. Running forecasts every \qty{48}{\hour} rather than every \qty{12}{\hour} is a pragmatic choice made to reduce computational cost and to keep the overall resource usage of the study within reasonable bounds. In the remainder of the paper, this configuration is referred to as the \emph{2025--48h} setup.

\subsection{Offline training experiments}

In this study, offline training is used both as a baseline approach and as a means to provide initial values for the model parameters in the subsequent online training experiments. While the overall method follows that of F25, some experimental choices adopted here differ slightly and are therefore described in detail.

\subsubsection{Training the hybrid model}

Instead of relying on operational analysis increments, we train the neural network offline using increments derived from a dedicated small reanalysis experiment. This reanalysis was produced using strong-constraint 4D-Var with IFS cycle 49r1, employing the same standard research configuration as described above, in particular a TCo399 resolution for the nonlinear model (the same resolution as the one used in the DA experiments). The reanalysis covers the period from April 2017 to March 2022. The first four years of data (April 2017 to March 2021) are used for training, while the final year (April 2021 to March 2022) is reserved for validation.

Using increments from this reanalysis dataset presents several advantages compared to operational increments. First, the forecast model is fully consistent throughout the entire dataset, whereas operational increments reflect the regular updates in model cycle over time. Second, the model configuration used in the reanalysis is identical to that employed in the online experiments, in particular in terms of horizontal resolution, which avoids potential inconsistencies between offline and online training stages. Finally, because the reanalysis is produced with strong-constraint 4D-Var, the full model error signal is contained in the analysis increments. In contrast, operational increments have been generated using weak-constraint 4D-Var since June 2020, in which case the model error signal is partly represented through diagnosed bias terms and is therefore split between multiple components.

From this reanalysis experiment, analysis states (inputs) and analysis increments (targets) are extracted at coarse resolution. Specifically, the fields are spectrally truncated at T15 and interpolated onto a $16 \times 32$ Gaussian grid. The neural network is then trained offline in prediction mode, using an MSE loss function, following exactly the same procedure as described in F25. The offline validation scores obtained with this setup are qualitatively and quantitatively very similar to those reported by F25. Since the primary focus of this paper is on the impact on forecast performance, results from the offline validation are not shown here.

\subsubsection{Forecasts from unseen data}
\label{sssec:offline-forecasts}

Once the neural network has been trained offline, it is embedded within the IFS to construct the hybrid forecast model. As described in \cref{ssec:methodology-offline}, the output of the neural network, originally trained to predict model error accumulated over one DA window, is rescaled to represent an effective correction at the model time-step scale and applied as an additive bias correction within the forecast model. The correction is not held constant over the full \qty{10}{\day} forecast period. Instead, it is updated every \qty{3}{\hour}. This update frequency was found to offer a good compromise between forecast accuracy and computational cost: while updating the correction at every model time step would be optimal in principle, it would also lead to a high increase in computational expense. In the current setup, enabling the NN correction increases the computational cost by less than \qty{3}{\percent}.

Using this hybrid model, we perform online forecast experiments following the 2025--48h setup described above. Importantly, the hybrid model is evaluated over the year 2025, which lies entirely outside the data used for both offline and online training. The results therefore provide an assessment of the generalisation capability of the offline-trained hybrid model.

\begin{figure}
    \centering
    \includegraphics[width=0.99\linewidth]{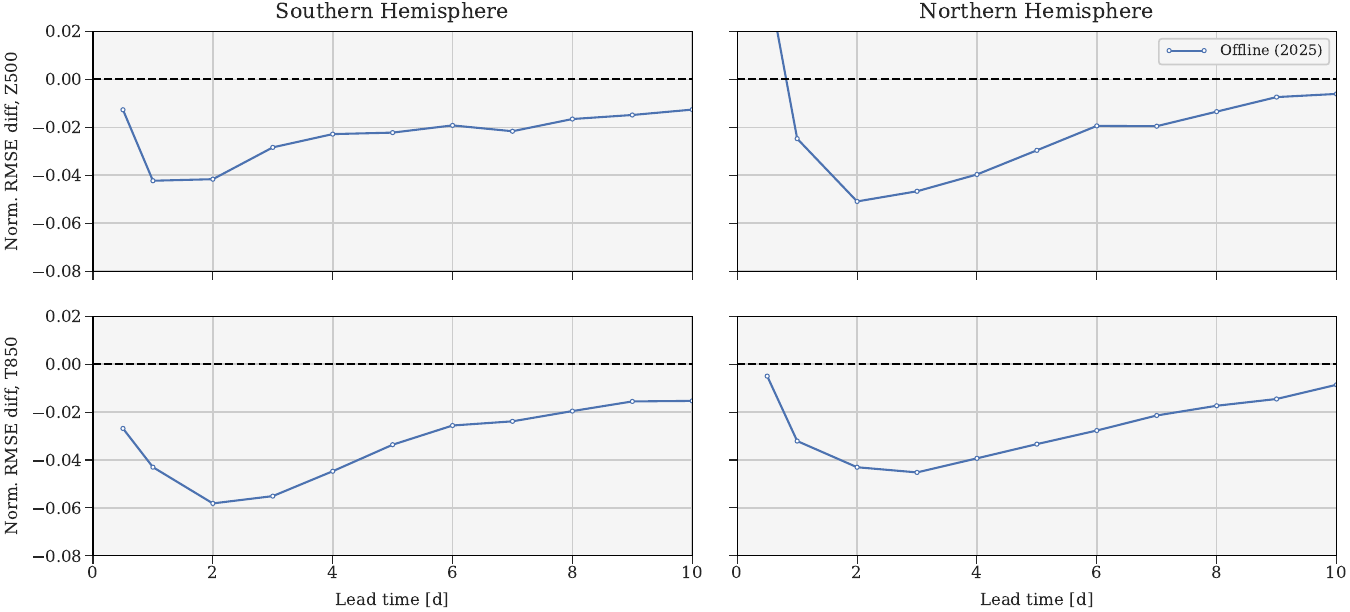}
    \caption{Normalised difference in RMSE between the forecasts of the offline-trained hybrid model and the control forecasts, as described in \cref{sssec:offline-forecasts}, for Z500 (top panels) and T850 (bottom panels) in the NH (right panels) and in the SH (left panels), evaluated over 2025.}
    \label{fig:results-offline}
\end{figure}

\Cref{fig:results-offline} shows the normalised difference in RMSE between the hybrid model forecasts and the control forecasts (i.e., the forecasts produced with the uncorrected model) for geopotential height at \qty{500}{\hecto\pascal} (Z500) and temperature at \qty{850}{\hecto\pascal} (T850). Scores are shown separately for the Northern Hemisphere (NH) and Southern Hemisphere (SH), and the operational analysis is used as verifying truth. Negative values correspond to an improvement of the hybrid model relative to the control.

Overall, these scores indicate that the offline-trained neural network has a positive impact on forecast accuracy. We get error reductions of up to approximately \qty{4}{\percent} not only for temperature, which is directly corrected by the neural network, but also for geopotential height, which is not explicitly corrected. This demonstrates that the correction interacts constructively with the physical model dynamics and induces improvements that propagate across variables.

However, a closer examination reveals that the magnitude of these improvements remains relatively modest. This is particularly noteworthy when compared to offline evaluation results, in which the same neural network is able to explain on the order of 25\% of the variance of the analysis increments for temperature, as measured by the relative MSE in F25. Although the present experimental setup slightly differs from F25, this order of magnitude remains representative.

In addition, the reduction in forecast errors peaks after approximately one to two days and then gradually vanishes at longer lead times. This behaviour suggests that, while the offline-trained correction is effective at compensating for short-term model error, it is not optimally aligned with the evolving model trajectories over extended forecast ranges.

Taken together, these results illustrate the intrinsic limitations of purely offline training strategies. While such approaches can provide useful initial corrections, they do not fully account for the interactions between the neural network correction and the physical forecast model. This highlights the need for online training methods, in which these interactions are explicitly incorporated into the learning process, as will be illustrated in the following sections.

\subsection{Sequential online training}

We now present a first set of three sequential online training experiments. The aim of these experiments is to illustrate both the strengths and limitations of the original online training framework used by F25 and of the targeted NN 4D-Var formulation proposed in the present work.

\subsubsection{Training experiments}
\label{sssec:seq-online-training-description}

In the first experiment, we employ the original NN 4D-Var method presented in \cref{ssec:methodology-nn-4d-var}, in which the initial state and the neural network parameters are estimated jointly within each assimilation window. Following F25, the background error covariance matrix for the model parameters is prescribed as $\mathbf{P} = \sigma_{\mathsf{p}}^{2}\mathbf{I}$, where $\mathbf{I}$ denotes the identity matrix and $\sigma_{\mathsf{p}} = \num{0.001}$. The interpretation and practical role of the parameter $\sigma_{\mathsf{p}}$ will be discussed in \cref{ssec:discussion-similarities-sgd}. Also note that the initial $\mathbf{p}^{\mathsf{b}}$ (i.e. for the very first assimilation cycle) corresponds to the parameters of the offline-trained NN. All other aspects of the experimental setup follow the standard research DA configuration used at ECMWF. In particular, the length of the assimilation window is set to \qty{12}{\hour}.

In the second experiment, we use the targeted NN 4D-Var method introduced in \cref{ssec:methodology-targ-nn-4d-var}, in which the assimilation process focuses exclusively on estimating the neural network parameters. In this case, the initial state is not part of the control vector and is prescribed using a precomputed analysis. Here, we use the operational analysis as initial state for each assimilation window. Apart from this change, all other experimental settings are identical to those of the first experiment, including the window length of \qty{12}{\hour}.

The third experiment is close to the second one, but differs in the length of the assimilation window, which is extended to \qty{48}{\hour}. In a standard cycled data assimilation setup, such long windows are typically impractical because the background trajectory rapidly diverges from the observations at longer lead times, primarily due to model error. Here, this difficulty is alleviated by the use of the operational analysis as initial state, which ensures that the initial trajectory is already close to the observations. In addition, the presence of an explicit model error correction mechanism further mitigates divergence issues over extended windows. From a machine learning perspective, increasing the window length provides more observations to constrain the model parameters and improves the signal-to-noise ratio, as model errors have more time to develop over \qty{48}{\hour} than over \qty{12}{\hour}.

All data assimilation experiments described in this subsection are run sequentially over the year 2024, with the first assimilation window starting on 2023-12-31 at 21:00 UTC. The experiments are run until 02 November 2024, for a total of \num{10} months of online training.

\subsubsection{Forecasts of the day}
\label{sssec:seq-online-fod}

After each assimilation cycle, a \qty{10}{\day} forecast is launched \qty{3}{\hour} into the assimilation window, corresponding to initial times at 00:00 UTC and 12:00 UTC. In the following paragraphs, we evaluate the quality of these forecasts, hereafter referred to as \emph{forecasts of the day}. The control forecasts are computed using the uncorrected model initialised from the operational analysis. This setup is identical to that used in the previous section, except that it covers the year 2024 instead of 2025. Furthermore, we include here two additional baselines.
\begin{itemize}
    \item The first baseline corresponds to the forecasts of the day in a data assimilation experiment similar to what is used in the control but running at TCo399 instead of TCo1279. The only purpose of this experiment is to show a typical accuracy that can be obtained when the forecast are started from an analysis computed at reduced horizontal resolution (namely at TCo399, as in all online training experiments) compared to the operational analysis (computed at TCo1279). In the following paragraph, this experiment will be referred to as the \emph{standard experiment}. 
    \item The second baseline corresponds to the forecasts produced with the offline-trained hybrid model, initialised from the operational analysis.
\end{itemize}
As an additional baseline, we also include forecasts produced with the offline-trained hybrid model, initialised from the operational analysis.

\begin{figure}
    \centering
    \includegraphics[width=0.99\linewidth]{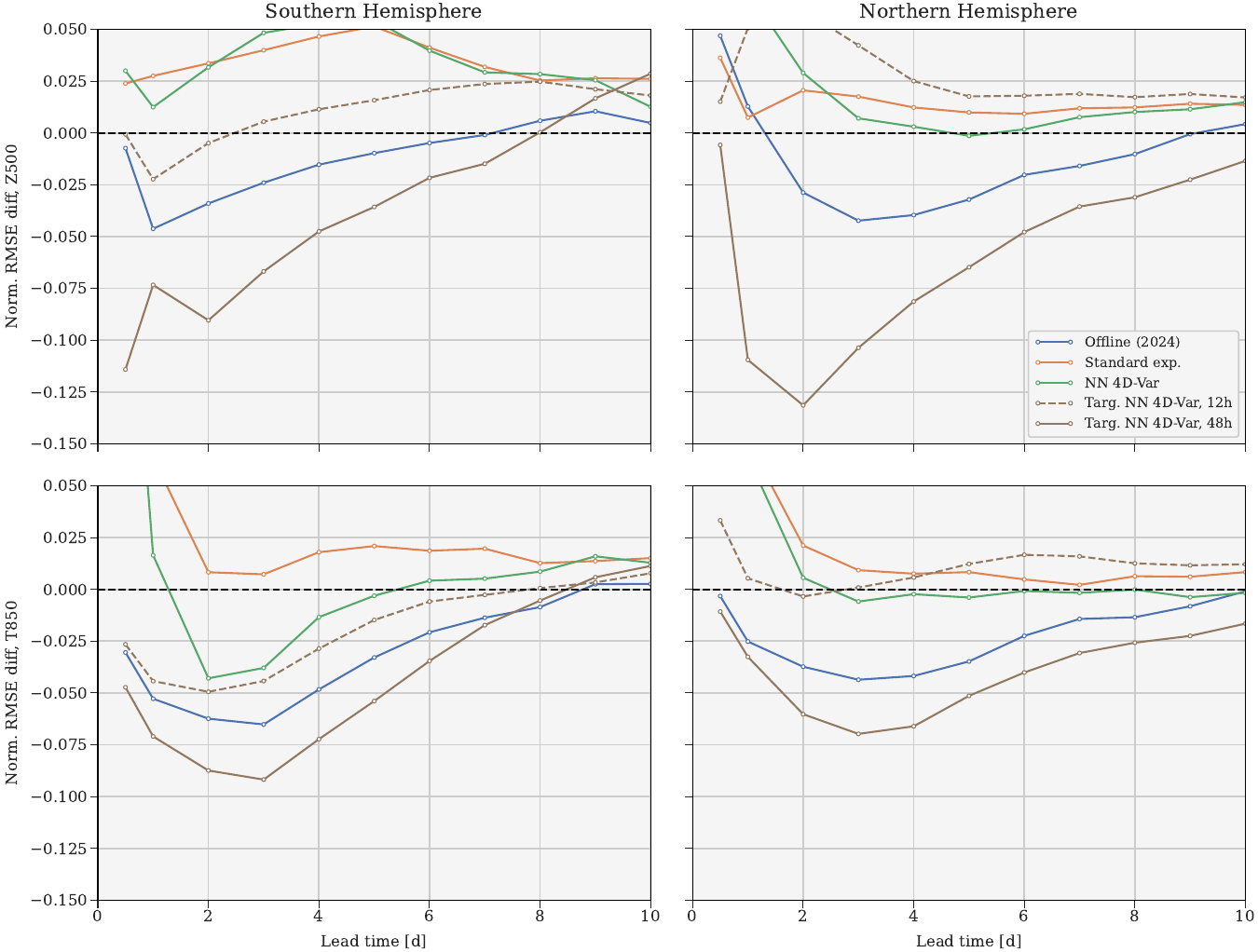}
    \caption{Same as \cref{fig:results-offline} for the forecasts of the day described in \cref{sssec:seq-online-fod}.}
    \label{fig:results-online-fod}
\end{figure}

\Cref{fig:results-online-fod} shows the normalised difference in RMSE for Z500 and T850. A first notable result is that the forecasts produced in the standard experiment are systematically less accurate than the control forecasts. This degradation is particularly pronounced at short lead times for temperature. These results show that the forecasts are in general more accurate when initialised from operations, therefore highlighting that, in the present comparison, the forecasts of the offline-trained hybrid model have a small advantage because they are initialised from the operational analysis.

Overall, the impact of the \qty{12}{\hour} window NN experiments (original NN 4D-Var and targeted NN with a \qty{12}{\hour} window) is mixed. In some cases, these forecasts are more accurate than those of the standard experiment (which does not benefit from a model error correction in the forecast), but in general it is clear that these forecasts are significantly less accurate than the forecasts of the offline-trained hybrid model. By contrast, when using the targeted NN 4D-Var formulation with a \qty{48}{\hour} window, forecast skill is improved in a consistent manner across variables and regions, with error reductions exceeding \qty{10}{\percent} for Z500 up to day \num{3}. This is the only configuration which visibly improves upon the offline-trained hybrid model.

While improvements in the forecasts of the day provide clear evidence that model error correction can be learned within the NN 4D-Var framework, these results must be interpreted with caution. First, the forecasts considered here do not correspond to a single, fixed hybrid model, but rather to a sequence of models: the neural network parameters are updated at each assimilation cycle, therefore the forecasts of the day are produced using a \enquote{model of the day}. Second, the \qty{48}{\hour} window experimental setup is not directly compatible with operations, as it relies on observations extending into the future of the forecast initial time. Although this limitation could in principle be alleviated (for instance by delaying the forecast initialisation within the assimilation window to match operational schedules) the primary objective of the present work is to design hybrid models that generalise across datasets and configurations. For this reason, in the following section, we focus on the evaluation of forecasts on unseen data.

\subsubsection{Forecasts from unseen data}
\label{sssec:seq-online-forecasts}

In order to evaluate the generalisation capabilities of the online-trained hybrid models, we regularly extract the NN parameters during the online training experiments (in 2024) and use them in online forecast experiments following the 2025--48h setup, exactly as in \cref{sssec:offline-forecasts}.

\begin{figure}
    \centering
    \includegraphics[width=0.99\linewidth]{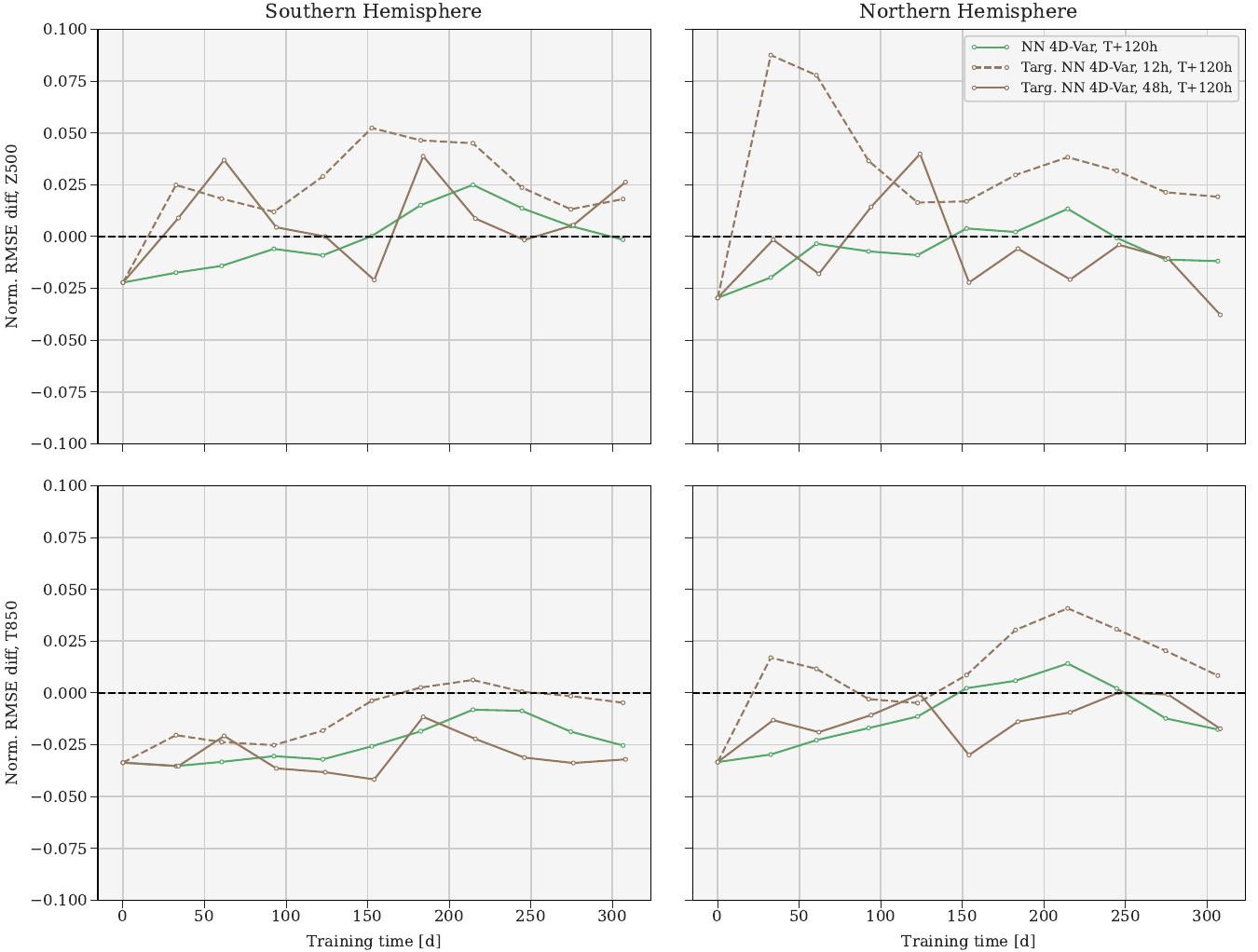}
    \caption{Normalised difference in RMSE at day \num{5} between (i) the forecasts of the hybrid models trained using sequential online training, as described in \cref{sssec:seq-online-forecasts}, and (ii) the control forecasts as a function of online training time (number of assimilation cycles multiplied by the length of the assimilation window), for Z500 (top panels) and T850 (bottom panels) in the NH (right panels) and in the SH (left panels), evaluated over 2025. As the NN are first pre-trained offline, the scores at training time zero exactly correspond to the offline trained hybrid shown in the previous figures.}
    \label{fig:results-seq-online-forecast}
\end{figure}

\Cref{fig:results-seq-online-forecast} shows the normalised difference in RMSE at day \num{5} between the hybrid model forecasts and the control forecasts for Z500 and T850  as a function of training time. The results are qualitatively very similar for all lead times throughout the \qty{10}{\day} of forecast (not illustrated here). The results are unambiguous and show that in the three experiments, the hybrid model progressively deteriorate over time. In other words, the information learned during the online training phase does not generalise to unseen data. This is particularly striking for the targeted NN 4D-Var experiment with a \qty{48}{\hour} window, for which the forecasts of the day showed substantial improvements, but whose performance collapses when evaluated in this out-of-sample forecast setting.

These results highlight fundamental limitations of the sequential online training framework and directly motivated the development of the batched training strategy illustrated in the following section, which aims to overcome these limitations and improve the generalisation capability of the hybrid model.

\subsection{Non-sequential online training}

We now present a second set of two online training experiments based on the non-sequential, batched framework. 

\subsubsection{Training experiments}

In both experiments, we employ the batched NN 4D-Var method introduced in \cref{ssec:methodology-batched-nn-4d-var}. As in the targeted NN 4D-Var formulation, the assimilation process focuses exclusively on estimating the neural network parameters, but, in contrast to the sequential framework, assimilation windows are processed in a random order.

As in the sequential experiments, the background error covariance matrix for the model parameters is set as $\mathbf{P} = \sigma_{\mathsf{p}}^{2}\mathbf{I}$, with $\sigma_{\mathsf{p}} = \num{0.001}$, and the initial $\mathbf{p}^{\mathsf{b}}$ corresponds to the parameters of the offline-trained NN.

The first experiment uses a batch size $N_{\mathsf{e}}=1$, while the second uses $N_{\mathsf{e}}=8$, in order to assess the sensitivity of the method to the batch size and to the associated mini-batch averaging. Both experiments rely on \qty{48}{\hour} assimilation windows. The dataset of windows used for training is identical to that used in the sequential online experiments, and consists of the \num{183} windows covering the year 2024 (without overlapping between assimilation windows).

Following standard practice in machine learning, we adopt an \emph{epoch}-based sampling strategy. At each optimisation step, $N_{\mathsf{e}}$ windows are randomly selected from the dataset without replacement. This process is repeated until all windows have been used once, which defines the completion of one epoch. A new epoch is then started by reshuffling the dataset and repeating the selection process.

Both experiments are run for a total of \num{183} optimisation steps. This corresponds to \num{1} epoch for the configuration with $N_{\mathsf{e}}=1$, and to \num{8} epochs for the configuration with $N_{\mathsf{e}}=8$, reflecting the increased number of windows processed per optimisation step in the batched setting.

\subsubsection{Forecasts from unseen data}
\label{sssec:rand-online-forecasts}

Following the experimental setup described in \cref{sssec:seq-online-forecasts}, we regularly extract the NN parameters during the online training experiments (in 2024) and use them in online forecast experiments following the 2025-48h setup.

\begin{figure}
    \centering
    \includegraphics[width=0.99\linewidth]{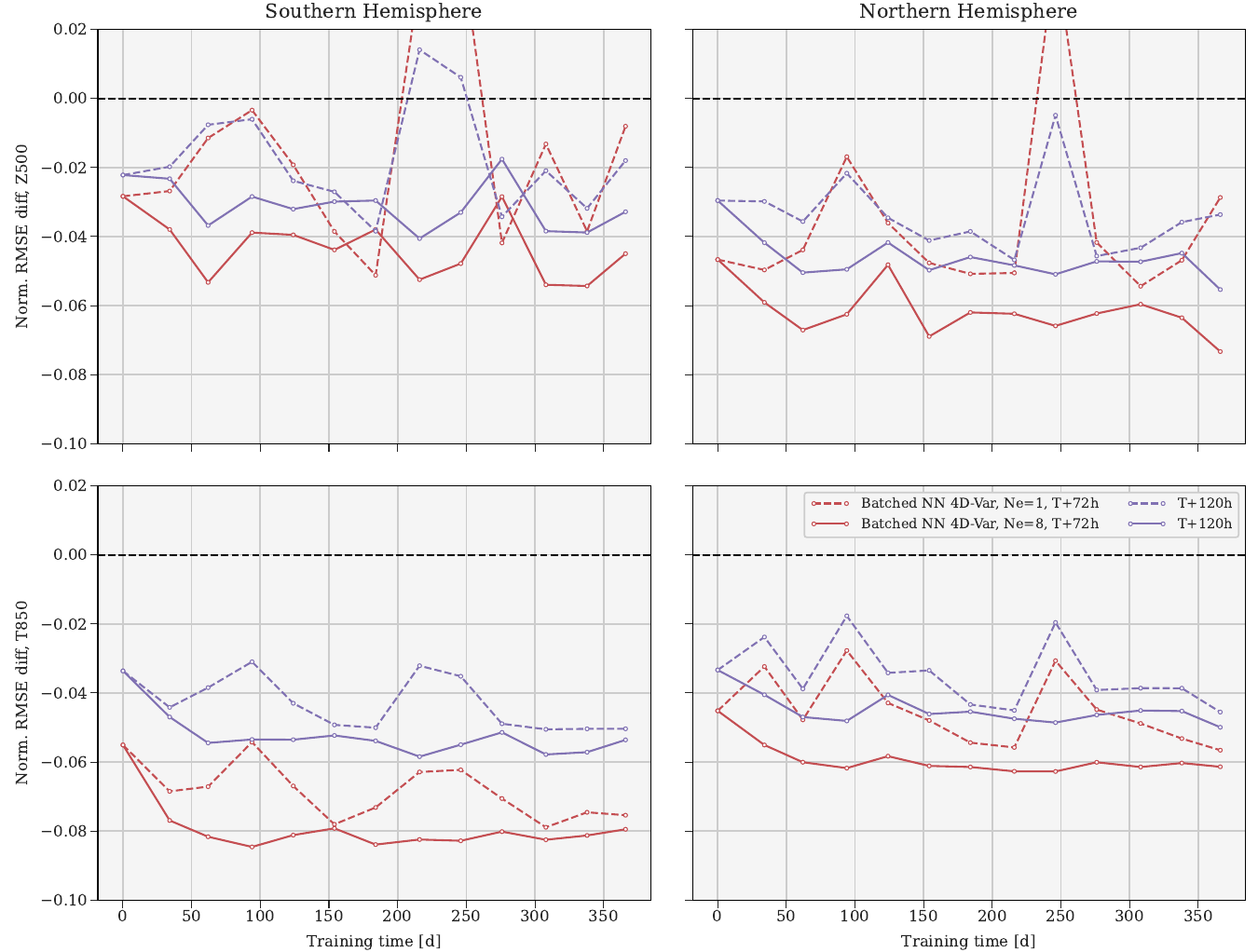}
    \caption{Same as \cref{fig:results-seq-online-forecast} at day \num{3} (in red) and at day \num{5} (in purple) for the hybrid models trained using non-sequential online training, as described in \cref{sssec:rand-online-forecasts}.}
    \label{fig:results-rand-online-forecast}
\end{figure}

\Cref{fig:results-rand-online-forecast} shows the normalised difference in RMSE at day \num{3} and at day \num{5} between the hybrid model forecasts and the control forecasts for Z500 and T850  as a function of training time. Contrary to the sequential training framework, the experiment with a batch size of $N_{\mathsf{e}}=8$ exhibits a clear improvement in forecast skill as training progresses. Although the convergence is not strictly monotonic, owing to the stochastic nature of the training process, the overall trend is unambiguously positive.

A similar behaviour is observed for the experiment with $N_{\mathsf{e}}=1$, particularly for temperature. However, in this configuration the signal is considerably noisier, making the convergence more difficult to identify. This behaviour is expected and can be attributed to two complementary effects. First, averaging parameter increments over \num{8} assimilation windows reduces the intrinsic noise of the updates and leads to a more stable optimisation process. Second, processing multiple windows at each optimisation step allows the algorithm to cycle through the training dataset more rapidly, effectively exposing the model to a larger number of samples for a given number of optimisation steps. Together, these effects explain the faster and more robust convergence obtained with larger batch sizes.

To further characterise the learning dynamics, we monitored the norm of the parameter increments throughout the training process (not shown). These diagnostics reveal an initial transition phase during which the magnitude of the parameter increments remains relatively large. This phase most likely corresponds to the adaptation of the neural network from a representation of analysis increments, inherited from the offline pre-training stage, toward a representation of instantaneous model error, which is the quantity required by the hybrid model formulation. Following this transition, which lasts approximately 50 to 100 days, the norm of the parameter increments stabilises at a level that is, in principle, controlled by the choice of $\sigma_{\mathsf{p}}$. Interestingly, \cref{fig:results-rand-online-forecast} indicates that the rate at which the forecast RMSE decreases also slows down after this transition. This behaviour suggests that the training could benefit from a progressive reduction of $\sigma_{\mathsf{p}}$ during the optimisation process, in a manner analogous to learning-rate scheduling in machine learning. This point is discussed further in \cref{ssec:discussion-similarities-sgd}. Alternatively, these results may indicate that a longer training dataset is required to continue improving the accuracy of the hybrid model.

\begin{figure}
    \centering
    \includegraphics[width=0.47\linewidth]{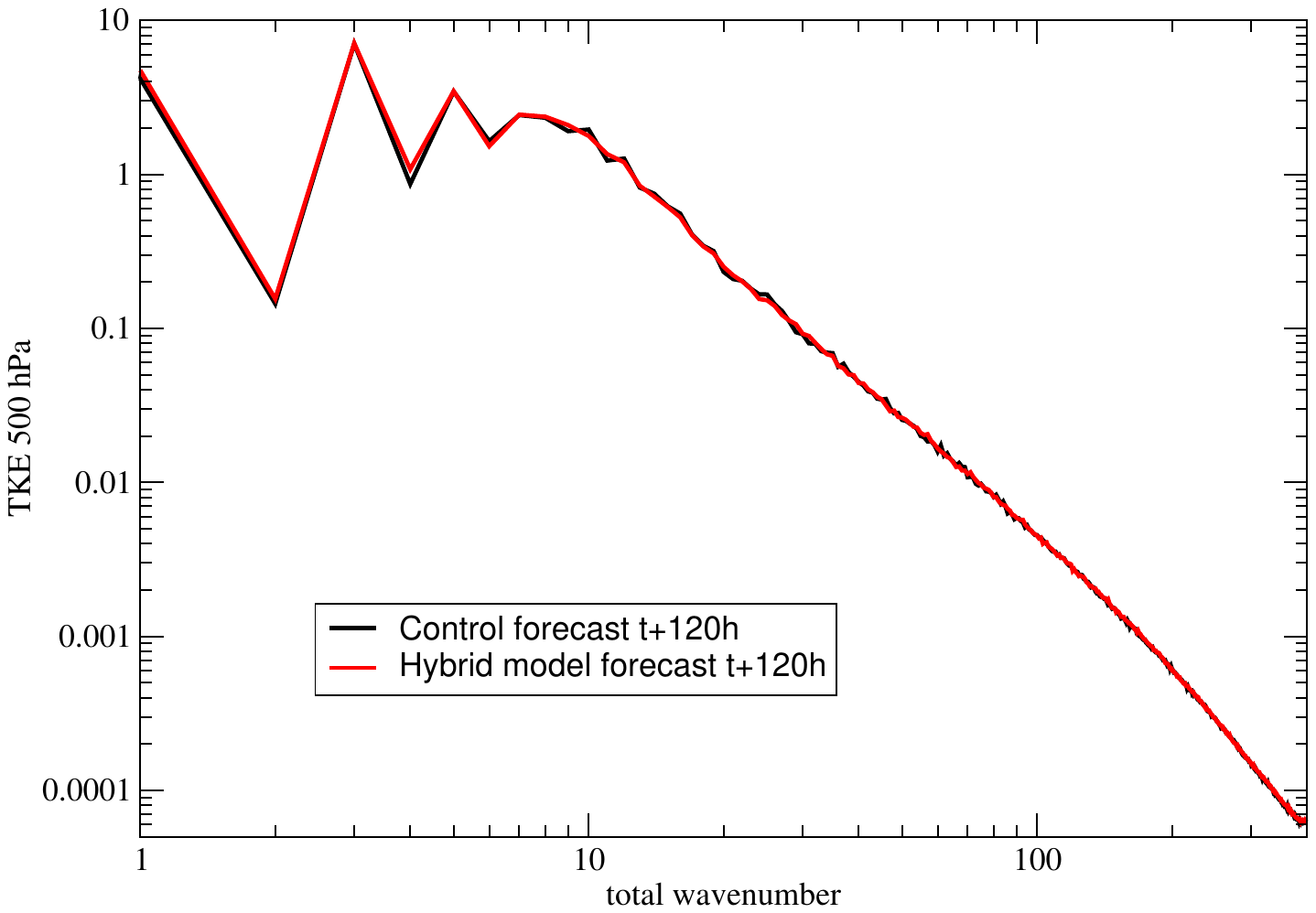} \includegraphics[width=0.47\linewidth]{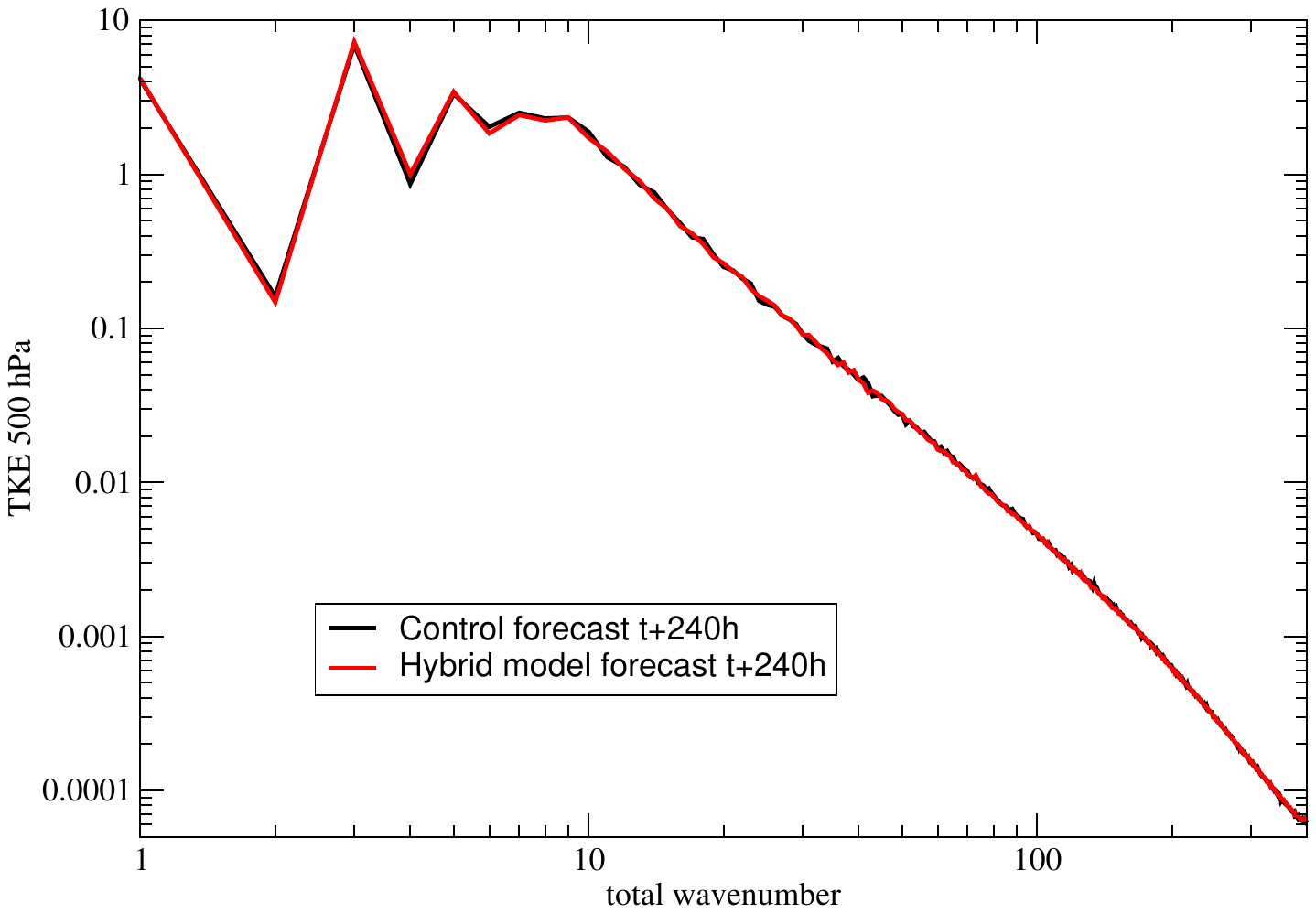} 
    \caption{Total kinetic energy for vector winds at \qty{500}{\hecto\pascal} for the control and hybrid forecasts at day \num{5} (left panel) and at day \num{10} (right panel), averaged over the first two months of forecast (January to February 2025).}
    \label{fig:results-rand-online-tke}
\end{figure}

Finally, we further analyse the hybrid model trained with $N_{\mathsf{e}}=8$ after \num{244} days (\num{8} months) of training, corresponding to \num{122} optimisation steps, or slightly more than five epochs. Two additional diagnostics are considered. First, \cref{fig:results-rand-online-tke} shows the total kinetic energy of the winds at \qty{500}{\hecto\pascal} for the hybrid model and control forecasts. The two distributions are visually indistinguishable, indicating that the forecast improvements obtained with the hybrid model are not achieved through an artificial smoothing of the flow. This is an important result, as many deterministic machine-learning weather prediction systems obtain part of their skill through a reduction of small-scale variability. By contrast, the present hybrid approach preserves realistic forecast dynamics while improving forecast accuracy.

\begin{figure}
    \centering
    \includegraphics[width=0.99\linewidth]{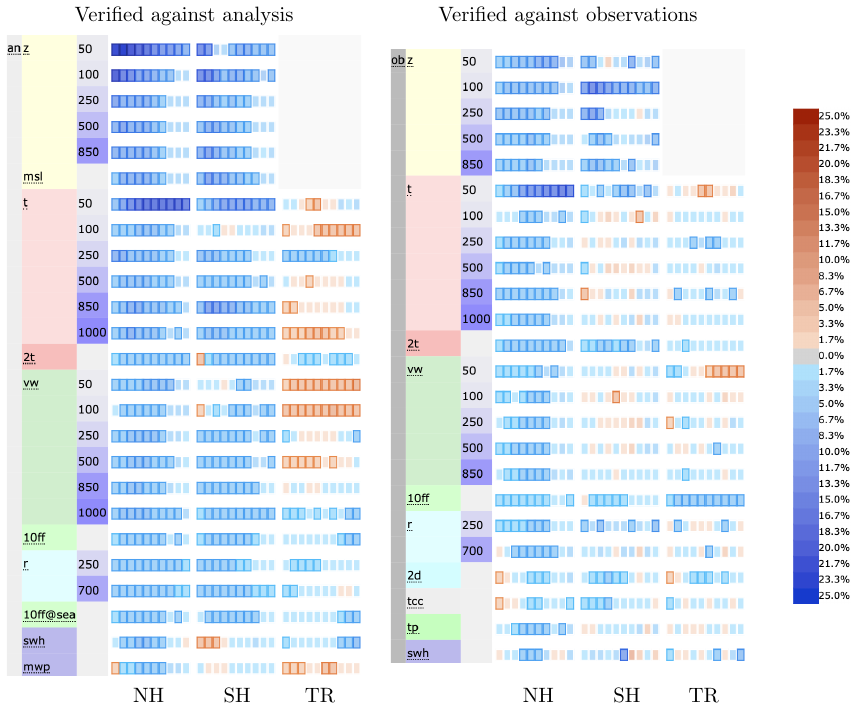}
    \caption{Summary scorecard comparing forecast RMSE differences between the hybrid model and the control forecasts in the 2025-48h setup, verified against operational analysis and against independent observations. Each box corresponds to a \qty{24}{\hour} lead-time interval from forecast day \num{1} to day \num{10}. The colour saturation is proportional to the magnitude of the normalised differences, with blue meaning improvement relative to control and red the other way around.}
    \label{fig:results-scorecard}
\end{figure}

Second, \cref{fig:results-scorecard} presents a summary scorecard of the hybrid-model performance. Overall, the scorecard confirms a positive impact of the online-trained correction across a wide range of variables and levels, notably in the stratosphere, where the hybrid model consistently reduces forecast errors relative to the control. At the same time, the scorecard highlights the main limitations of the current approach, with the most significant degradations remaining concentrated in the tropics. These results therefore provide both a validation of the proposed training framework and a clear indication of the scientific priorities for future developments.

\section{Discussion}
\label{sec:discussion}

We now discuss some methodological aspects of our framework and of the optimisation implementation.

\subsection{Coordinate descent interpretation}
\label{ssec:discussion-coordinate-descent}

Hybrid machine learning–data assimilation (ML--DA) approaches designed to identify the dynamics of a system from sparse and noisy observations are often grounded in a general variational formulation \citep{abarbanel-2018, bocquet-2019a, bocquet-2020} derived from Bayesian principles. Under standard Gaussian error assumptions, this framework leads to the following weak-constraint 4D-Var cost function:
\begin{align}
    \mathcal{J}\left(\mathbf{p}, \mathbf{x}_{0}, \ldots, \mathbf{x}_{N_{\mathsf{t}}}\right) ={}& \frac{1}{2}\left\|\mathbf{x}_{0}-\mathbf{x}^{\mathsf{b}}_{0}\right\|^{2}_{\mathbf{B}^{-1}} + \frac{1}{2}\left\|\mathbf{p}-\mathbf{p}^{\mathsf{b}}\right\|^{2}_{\mathbf{P}^{-1}} + \frac{1}{2}\sum_{k=0}^{N_{\mathsf{t}}}\left\|\mathbf{y}_{k} - \boldsymbol{\mathcal{H}}_{k}\left(\mathbf{x}_{k}\right)\right\|^{2}_{\mathbf{R}^{-1}_{k}} \nonumber \\ &+ \frac{1}{2}\sum_{k=0}^{N_{\mathsf{t}}-1}\left\|\mathbf{x}_{k+1} - \boldsymbol{\mathcal{M}}^{\mathsf{h}}_{k+1:k}\left(\mathbf{p}, \mathbf{x}_{k}\right)\right\|^{2}_{\mathbf{Q}^{-1}_{k}},
\end{align}
where $\mathbf{Q}$ denotes the model error covariance matrix and $N_{\mathsf{t}}$ the length of the observation dataset. Minimising this cost function means estimating both the system trajectory $\mathbf{x}_{0}, \ldots, \mathbf{x}_{N_{\mathsf{t}}}$ and the model parameters $\mathbf{p}$.

In realistic applications, the dataset length $N_{\mathsf{t}}$ must be sufficiently large in order to capture the relevant dynamics of the system. As a consequence, direct joint optimisation over both the trajectory and the model parameters quickly becomes computationally intractable. To address this issue, \citet{bocquet-2020, brajard-2020} proposed an iterative coordinate descent strategy, in which the estimation of the system trajectory and that of the model parameters are decoupled. In this framework, DA methods are used to estimate the trajectory, while machine learning techniques are used to identify the model parameters, the two steps being alternated iteratively.

The offline training strategy introduced by \citet{farchi-2021} fits naturally within this DA/ML iterative framework. Starting from an initial guess for the model parameters (typically set to zero, so that the hybrid model reduces to the original physical model), a DA step is first applied to estimate the system trajectory. For large datasets, this is achieved using cycled DA methods, which effectively partition the observations into consecutive assimilation windows, yielding a (re)analysis dataset $\mathbf{x}^{\mathsf{a}}_{0}, \ldots, \mathbf{x}^{\mathsf{a}}_{N_{\mathsf{t}}}$. In a second step, machine learning techniques are used to update the model parameters based on this dataset; in the present case, the neural network is trained to predict analysis increments. This completes one iteration of the DA/ML loop.

In principle, the updated hybrid model could then be used to recompute a new (re)analysis dataset, leading to further refinement of the trajectory and then of the model parameters \citep{bocquet-2020, brajard-2020, bocquet-2023}. However, in the context of model error correction, with existing data assimilation systems that are typically already well calibrated, the computational cost of recomputing a full (re)analysis dataset is usually not justified by the marginal gains in accuracy. As a result, practical implementations commonly stop after a single DA/ML iteration.

The NN 4D-Var online training method introduced by \citet{farchi-2021b} departs from this iterative framework by seeking to estimate the model parameters and the system trajectory simultaneously within a single variational problem. In order to remain computationally tractable, the observation dataset is partitioned into consecutive assimilation windows and the optimisation is performed sequentially.

The targeted NN 4D-Var and batched NN 4D-Var formulations introduced in the present work can be interpreted as a return to the DA/ML iterative framework. In both cases, the estimation of model parameters is performed separately, assuming that a precomputed analysis is available to describe the system trajectory. As a consequence, while these approaches depart from the traditional sequential DA framework, they remain consistent with the underlying variational formulation and retain a clear connection to the established DA/ML iterative methodology.

\subsection{Simplifying the targeted NN 4D-Var method}
\label{ssec:discussion-gradient-formulation}

In this section, we would like to show how the targeted NN 4D-Var method can be simplified and highlight its connections with the stochastic gradient descent method widely used in ML. To do so, let us consider the following linear 3D-Var problem:
\begin{equation}
    \mathcal{J}(\mathbf{x}) = \frac{1}{2}\left\|\mathbf{x}-\mathbf{x}^{\mathsf{b}}\right\|^{2}_{\mathbf{B}^{-1}} + \frac{1}{2}\left\|\mathbf{y} - \mathbf{H}\mathbf{x}\right\|^{2}_{\mathbf{R}^{-1}},
\end{equation}
where $\mathbf{H}$ is a linear observation operator, and where we assume $\mathbf{B} = \sigma^2 \mathbf{I}$. In this setting, the gradient of the cost function evaluated at $\mathbf{x} = \mathbf{x}^{\mathsf{b}}$, often referred to as the \emph{first-guess gradient}, is given by
\begin{equation}
    \boldsymbol{\nabla}_{\mathbf{x}}\mathcal{J}\left(\mathbf{x}^{\mathsf{b}}\right) = -\mathbf{H}^{\top}\mathbf{R}^{-1}\mathbf{d},
\end{equation}
with $\mathbf{d}\triangleq\mathbf{y}-\mathbf{Hx}^{\mathsf{b}}$, and the analysis increment is given by
\begin{equation}
    \delta\mathbf{x} = \sigma^2\mathbf{H}^{\top}\left(\mathbf{R}+\sigma^2\mathbf{HH}^{\top}\right)^{-1}\mathbf{d}.
\end{equation}
In the regime where $\sigma^2$ is small compared to the observation error variances, this simplifies into
\begin{equation}
\label{eq:4d-var-an-inc}
    \delta\mathbf{x} \approx \sigma^2\mathbf{H}^{\top}\mathbf{R}^{-1}\mathbf{d} = -\sigma^2 \boldsymbol{\nabla}_{\mathbf{x}}\mathcal{J}\left(\mathbf{x}^{\mathsf{b}}\right),
\end{equation}
showing that the increment is proportional to $\sigma^2$.

Coming back to our targeted NN 4D-Var problem, assuming $\mathbf{B} = \sigma_{\mathsf{p}}^2 \mathbf{I}$ (as we have used in the numerical experiments), \cref{eq:4d-var-an-inc} becomes
\begin{equation}
\label{eq:nn-4d-var-pan-inc}
    \delta\mathbf{p} \approx -\sigma_{\mathbf{p}}^2 \boldsymbol{\nabla}_{\mathbf{p}}\mathcal{J}\left(\mathbf{p}^{\mathsf{b}}\right),
\end{equation}
where $\boldsymbol{\nabla}_{\mathbf{p}}\mathcal{J}\left(\mathbf{p}^{\mathsf{b}}\right)$ denotes the gradient of the cost function with respect to the model parameters, evaluated at the background parameters. If $\sigma^2_{\mathsf{p}}$ is small compared to the observation error variances (in the numerical experiments, we have used $\sigma^2_{\mathsf{p}}=0.001$), we expect this approximation to remain valid at first order, even though the 4D-Var case introduces additional complexity due to temporal evolution and nonlinearities. Under this approximation, the targeted NN 4D-Var method becomes formally equivalent to stochastic gradient descent, with $\sigma_{\mathbf{p}}^2$ playing the role of the equivalent learning rate.

Building on this analogy, one may consider a simplified version of the targeted NN 4D-Var method in which the costly minimisation of the variational cost function, \cref{eq:targ-nn-4d-var-cost}, is replaced by a single evaluation of the gradient with respect to the model parameters. In this case, the parameter update would be given by \cref{eq:nn-4d-var-pan-inc}, without approximation. From a computational perspective, such a simplification would be highly attractive, as it would replace the cost of solving an incremental 4D-Var problem, typically involving multiple outer loops, by a single gradient calculation. This could lead to a substantial reduction in computational cost. However, it remains unclear whether such a simplified approach would be robust in practice. In particular, because the neural network correction is applied repeatedly throughout the assimilation window, the resulting gradients may exhibit strong amplification effects, potentially leading to unstable or exploding updates. In variational data assimilation, such issues are mitigated through the use of background constraints, for instance by prescribing relatively small error covariance matrices (e.g., $\mathbf{B}$ or $\mathbf{P}$), which effectively regularise the optimisation problem. By contrast, the tools available in standard machine learning for handling such instabilities are more limited, with gradient clipping being the main practical option. Further investigation is therefore required to assess whether this simplified, gradient-based formulation can be made stable and effective in practice, and to determine under which conditions it may provide a viable alternative to the full variational approach.

\subsection{Similarities with stochastic gradient descent}
\label{ssec:discussion-similarities-sgd}

As shown in the previous section, in the regime where $\sigma^2_{\mathsf{p}}$ is small compared to the observation error variances, the targeted NN 4D-Var formulation exhibits strong similarities with the stochastic gradient descent method widely used in ML. Building on this analogy, it is natural to consider incorporating additional strategies from stochastic optimisation in order to further improve the efficiency and robustness of the online training framework.

A first example is the introduction of momentum in the parameter update. In the current formulation, the update of the model parameters at cycle $t$ is given by
\begin{equation}
    \mathbf{p}^{\mathsf{a}}(t) = \mathbf{p}^{\mathsf{b}}(t) + \frac{1}{N_{\mathsf{e}}}\sum_{i=1}^{N_{\mathsf{e}}}\delta\mathbf{p}^{\mathsf{a}}_{i}(t).
\end{equation}
Introducing momentum would lead to the modified update
\begin{subequations}
    \begin{align}
        \Delta\mathbf{p}(0) &= 0,\\
        \Delta\mathbf{p}(t) &= \frac{1-\mu}{N_{\mathsf{e}}}\sum_{i=1}^{N_{\mathsf{e}}}\delta\mathbf{p}^{\mathsf{a}}_{i}(t) + \mu\Delta\mathbf{p}(t-1),\\
        \mathbf{p}^{\mathsf{a}}(t) &= \mathbf{p}^{\mathsf{b}}(t) + \Delta\mathbf{p}(t),
    \end{align}
\end{subequations}
where $\mu \in [0,1)$ controls the strength of the temporal smoothing. The case $\mu = 0$ corresponds to the original formulation, while $\mu > 0$ introduces a memory effect in the parameter updates. In the context of stochastic optimisation, momentum is commonly used to mitigate oscillations caused by noisy gradients and to accelerate convergence. A similar effect is expected here, where random variations arise from the sampling of assimilation windows.

More generally, the parameter update can be expressed in terms of an abstract optimisation operator,
\begin{subequations}
    \begin{align}
        \Delta\mathbf{p}(t) &= \mathcal{A}_t\left(\frac{1}{N_{\mathsf{e}}}\sum_{i=1}^{N_{\mathsf{e}}}\delta\mathbf{p}^{\mathsf{a}}_{i}(t)\right),\\
        \mathbf{p}^{\mathsf{a}}(t) &= \mathbf{p}^{\mathsf{b}}(t) + \Delta\mathbf{p}(t).
    \end{align}
\end{subequations}
where $\mathcal{A}_{t}(\cdot)$ denotes a generic optimisation algorithm, such as Adam \citep{kingma-2015} or related adaptive methods. This formulation highlights that the batched targeted NN 4D-Var framework can naturally accommodate more sophisticated update rules, potentially improving convergence speed and robustness.

A second avenue for improvement is the introduction of a learning rate scheduler. We have shown in the previous section that $\sigma^2_{\mathsf{p}}$ can be interpreted as the effective learning rate in our method. As suggested by the results of \cref{sssec:rand-online-forecasts}, the batched training experiments could benefit from a reduction of $\sigma_{\mathsf{p}}$ during the later stages of training. In machine learning, this idea is formalised through learning rate schedulers, in which the learning rate is progressively decreased as optimisation proceeds. Introducing a similar mechanism in the present framework would provide additional control over the convergence of the model parameters and could further improve the generalisation properties of the resulting hybrid models.

\section{Conclusions}
\label{sec:conclusions}

Hybrid modelling offers a promising route to combine the strengths of physics-based models with data-driven corrections. In this paper, we investigated how neural networks can be trained and deployed as physically consistent model-error corrections within an operational-style NWP model.

In our hybrid model formulation, a neural network provides an additive, state-dependent correction applied within the numerical time-integration scheme. Using this setting, we first implemented offline training based on analysis increments, employed here both as a baseline and as a pre-training step. Offline-trained corrections were shown to yield consistent but modest improvements in out-of-sample forecast experiments. These improvements peak at short lead times and then progressively diminish, highlighting a fundamental limitation of offline strategies: the learned correction does not take into account interactions with the physical model.

To overcome these issues, we revisited online training approaches based on the NN 4D-Var framework. We introduced a first methodological extension, \emph{targeted NN~4D-Var}, which removes joint state-parameter estimation by relying on precomputed analyses for the initial state and re-assimilating observations to estimate model parameters only. This reduces the risk of aliasing initial-condition errors into parameter updates and alleviates the need for difficult calibration of state-parameter background-error balances. However, sequential online learning experiments revealed a critical issue: although forecasts of the day could be visibly improved, the resulting online-trained corrections failed to generalise to unseen data.

Motivated by this finding, we proposed a second methodological contribution, \emph{batched NN 4D-Var}. By enabling windows to be processed in random order and by averaging parameter increments across mini-batches of windows, this approach directly addresses the limitations of sequential training. The batched formulation builds on the close analogy between targeted NN 4D-Var cycles and stochastic gradient descent. Results from out-of-sample forecast experiments demonstrated that batched training removes the generalisation failure observed in sequential online learning and yields hybrid model progressively improving as training proceeds. Furthermore, the hybrid model forecasts were found to be dynamically indistinguishable from the control forecasts.

Several directions of development emerge naturally from this work. The analogy with stochastic optimisation suggests that further gains may be achieved by incorporating additional tools from machine learning into the batched targeted NN 4D-Var framework, such as momentum, adaptive update rules, and learning-rate scheduling. A more radical simplification (replacing the inner variational minimisation by a direct first-guess-gradient update) could dramatically reduce computational cost, but would require careful investigation of stability.

A key practical limitation of the current study is that the correction had to be disabled in the tropics, where applying it was found to degrade forecast skill. Understanding and resolving this behaviour is a priority for future work. We hypothesise that (i) explicitly incorporating specific humidity into the neural-network inputs and/or outputs may be essential in tropical regimes, (ii) increasing the update frequency of the correction (e.g., from \qty{3}{\hour} to \qty{1}{\hour}) could improve the representation of diurnal and rapidly evolving processes, and (iii) moving beyond purely column-based architectures toward horizontally extended corrections (though not yet technically available in our current implementation) may be necessary to capture organised tropical structures and non-local interactions.

Overall, the proposed targeted and batched NN 4D-Var formulations provide a practically effective path toward generalisable hybrid model-error correction in NWP. By including non-sequential and mini-batch training strategies in a robust variational framework for online learning of model parameters, this work advances the design of hybrid ML--DA systems that improve forecast skill while retaining the dynamical characteristics of the underlying physics-based model. From an operational perspective, the approach naturally benefits from future improvements in observations, data assimilation, and the physical forecast model, since the model-error correction is learned directly within the existing forecasting system. Although model error is inherently model-dependent and is therefore expected to evolve with successive model upgrades, we expect these changes to remain sufficiently gradual that previously trained corrections can be reused as a starting point and incrementally fine-tuned. This suggests that the operational maintenance cost of such a hybrid system will remain limited, with retraining largely taking the form of periodic adaptation to new model cycles rather than repeated training from scratch. As such, providing performance can continue to be advanced (e.g. in the tropics) to the required level, we see no logistical limitations that would preclude this approach being considered in the future as a candidate for operational deployment.

\section*{Acknowledgements}

AF would like to thank Elias Holm for his help in setting up forecast experiments. The authors are grateful to Dr. T. McNally and Dr. I. Sandu for comments and suggestions on the manuscript. CEREA is a member of Institut Pierre--Simon Laplace.

\bibliography{bibtex}

\end{document}